# Stacking-Enhanced Bagging Ensemble Learning for Breast Cancer Classification with CNN

1st Peicheng Wu
*Electronics and Information Engineering*
*Huazhong University of Science and Technology*
Wuhan, China
pcwu9424@gmail.com

2nd Runze Ma
*Automation*
*Huazhong University of Science and Technology*
Wuhan, China
runzenevergiveup@foxmail.com

3rd Teoh Teik Toe
*NTU Business AI Lab*
*Nanyang Technological University*
Singapore
ttteoh@ntu.edu.sg

*Abstract*—This paper proposes a CNN classification network based on Bagging and stacking ensemble learning methods for breast cancer classification. The model was trained and tested on the public dataset of DDSM. The model is capable of fast and accurate classification of input images. According to our research results, for binary classification (presence or absence of breast cancer), the accuracy reached 98.84%, and for five-class classification, the accuracy reached 98.34%. The model also achieved a micro-average recall rate of 94.80% and an F1 score of 94.19%. In comparative experiments, we compared the effects of different values of bagging_ratio and n_models on the model, as well as several methods for ensemble bagging models. Furthermore, under the same parameter settings, our BSECNN outperformed VGG16 and ResNet-50 in terms of accuracy by 8.22% and 6.33% respectively.

*Keywords—breast cancer classification, CNN, bagging, stacking, VGG16, ResNet-50*

## I. INTRODUCTION

Breast cancer is a prevalent and fatal cancer that affects women globally. According to estimates [1], more than 2 million new cases of breast cancer were diagnosed in 2020 alone. Detecting breast cancer with precision and timeliness is crucial for enhancing the survival rates and overall well-being of patients [2]. Breast cancer examination results can generally be categorized into five groups: negative, benign calcification, benign mass, malignant calcification, and malignant mass. However, the classification of these results is a difficult undertaking, primarily due to the complexity, heterogeneity, and high dimensionality of the data involved [3].

Breast cancer classification aims to accurately diagnose the disease and predict tumor behavior, enabling informed decision-making in oncology. However, traditional methods relying on clinicopathological features and routine biomarker assessments may not fully capture the diverse clinical courses of individual breast cancers due to human errors or noise in the data [4]. To address this issue, two primary approaches are used. The first approach is molecular classification, which identifies different subtypes of breast cancer with distinct prognoses and treatment responses [5]. The second approach involves employing ensemble learning techniques to learn additional features and mitigate the impact of noise on the classification model.

Machine learning, a subset of AI, enables systems to learn from data and make predictions or decisions. The paper presents a new method for breast cancer classification that utilizes the Bagging and Stacking Ensemble Convolutional Neural Network (BSECNN). BSECNN is a hybrid approach that combines the advantages of ensemble learning methods such as Bagging and Stacking, with those of Convolutional Neural Network (CNN). Bagging is a technique that generates multiple models from bootstrap samples of the data and aggregates their predictions to reduce variance and improve generalization ability[6].

In this study, we utilized a dataset known as DDSM Mammography, which combines images from the DDSM and CBIS-DDSM datasets [7] [8]. Unlike previous research that focused on classifying pre-identified lesions, our dataset was specifically curated to classify raw scans as either positive or negative by detecting abnormalities. The automated detection of lesions has the potential to greatly contribute to saving numerous lives.

The dataset comprises a total of 55,890 training examples, with 14% classified as positive and the remaining 86% as negative. It includes negative images sourced from the DDSM dataset and positive images sourced from the CBIS-DDSM dataset. To process the negative (DDSM) images, they were divided into 598x598 tiles and subsequently resized to 299x299. On the other hand, for the positive (CBIS-DDSM) images, their regions of interest (ROIs) were extracted using masks, with a small amount of padding added for context. Each ROI was randomly cropped three times, resulting in 598x598 images with random flips and rotations. Finally, these images were resized to 299x299. The main contributions and novelties of this paper are as follows:

- We propose a novel breast cancer classification method based on a Bagging and stacking ensemble CNN network (BSECNN).

- We conducted extensive experiments on a publicly available dataset and investigated the selection of bagging_ratio and n_models to reduce training costs.

- We performed a comprehensive analysis of the performance and characteristics of our method, comparing it with other approaches.

## II. METHODOLGY

### A. Convolutional Neural Network

Convolutional Neural Networks (CNNs) are a class of feedforward neural networks that excel in processing large images. Unlike traditional neural networks, CNNs employ specialized neurons that selectively respond to specific patterns within a local receptive field. This allows them to efficiently extract features and capture spatial relationships in the input data, making them particularly well-suited for image-related tasks. CNNs have demonstrated impressive performance in various image processing applications.

CNNs have been shown to outperform other deep learning architectures in image and speech recognition tasks. This model can be trained using backpropagation algorithms to optimize the weights and biases of the network. Compared to other deep feedforward neural networks, CNNs require fewer parameters to achieve good performance, making them an attractive option for deep learning applications.

#### 1) Structure

A CNN (Convolutional Neural Network) is structured with different layers, including the input layer, convolutional layer, pooling layer, fully connected layer, and output layer.

The input layer is responsible for receiving the data, typically in the form of a three-dimensional matrix that represents the image's height, width, and number of channels.

The convolutional layer applies convolution operations to the input data using convolution kernels. These small matrices slide over the input data, extracting features and generating new feature maps. Multiple convolution kernels are typically used in this layer to extract different features.

The pooling layer is responsible for reducing the dimensionality of the feature maps obtained from the convolutional layer. It helps reduce computation and the number of parameters while also preventing overfitting.

The fully connected layer, also known as the dense layer, plays a crucial role in the convolutional neural network (CNN) architecture. It serves as the final stage of feature extraction and performs important operations such as classification or regression.

#### 2) Computation Process

The forward propagation process of a Convolutional Neural Network (CNN) is as follows: for input data X, firstly, the convolution operation is applied to obtain the convolutional feature map:

$$I = C(X, W) + b \quad (1)$$

where C represents the convolution operation, W is the convolutional kernel (weight matrix), and b is the bias term. The convolution operation can be defined as:

$$C(I, W) = \sum_{i=1}^{m} \sum_{j=1}^{n} I(i,j) * W(i,j) \quad (2)$$

Here, $C(I, W)$ denotes the result of the convolution operation, $I(i,j)$ represents the element at the $i-th$ row and $j-th$ column of the input feature map I, and $W(i,j)$ represents the element at the i-th row and $j-th$ column of the convolutional kernel $W$.

Next, the convolutional feature map $I$ is passed through an activation function to obtain the activation feature map $A = f(I)$, where f is the activation function. Subsequently, the activation feature map $A$ is downsampled using a pooling operation to obtain the pooled feature map $O = P(A)$, where $P$ represents the pooling operation. Finally, the pooled feature map O is input to a fully connected layer for linear transformation, resulting in the output vector:

$$Y = W' * O + b' \quad (3)$$

where W' is the weight matrix of the fully connected layer and b' is the bias term. The overall process can be represented as:

$$Y = W' * P\big(f(C(X,W) + b)\big) + b' \quad (4)$$

During backpropagation, the gradient of the output vector $Y(dY)$ is computed with respect to the loss function. This gradient is then successively back propagated through the layers, computing gradients for the weights and biases in each layer. These gradients are computed using the chain rule and are used to update the parameters of the network for optimization.

### B. Bagging Ensemble Methody

Bagging, which stands for Bootstrap Aggregating, is a powerful ensemble learning technique that is known for its effectiveness in bolstering the overall predictive performance. [10] This is achieved by aggregating the predictions derived from an array of distinct base learners. The essence of Bagging resides in its unique strategy of generating multiple training subsets; this is executed by carrying out bootstrap sampling [11] on the original dataset.

Bootstrap sampling is a method of sampling with replacement, which means that in each sampling iteration, some samples may be selected multiple times, while some samples may not be selected at all. Specifically, for a data set of size $N$, the probability of selecting a sample in each sampling iteration is $1/N$. Therefore, the probability that a sample is not selected in one sampling iteration is $1 - 1/N$. After $N$ sampling iterations (i.e., the process of bootstrap sampling), the probability that a sample is never selected is $(1 - 1/N)^N$. As N approaches infinity, this probability approaches $1/e$, which is approximately 0.368, or in other words, there is approximately a 36.8% chance that a sample will not be selected at all in bootstrap sampling. As a result, each submodel learns different features of the data set, leading to higher accuracy when the predictions are aggregated.

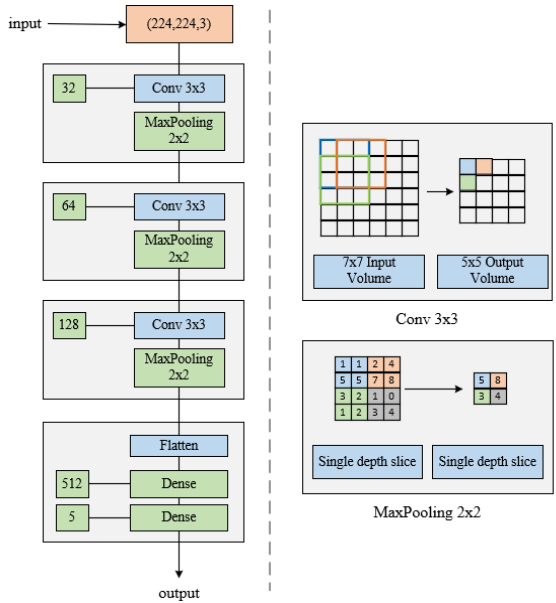

Fig. 1. The structure of the CNN in this paper

*C. Stacking Ensemble Methody*

Stacking is an ensemble learning method that aims to enhance the predictive performance of machine learning models. Introduced by Wolpert in 1992 [12], it addresses the limitations of individual models by leveraging the concept of combining their predictions. The fundamental idea behind Stacking is to integrate multiple diverse base models, such as decision trees, support vector machines, and neural networks, to obtain more accurate and robust predictions.

In Stacking, the training process involves dividing the dataset into multiple folds. Each fold is used to train different base models, generating individual predictions. These predictions are then utilized as new features to train a meta-model, typically a classifier or regressor. The meta-model learns how to effectively combine the predictions of the base models to produce the final prediction outcome. In our experiment, we made some modifications by using bagging with bootstrap sampling to train models that learned different features as sub-models. This approach has improved both the accuracy and precision of the overall model.

One of the key advantages of Stacking is its ability to harness the strengths of various models, leading to improved overall predictive performance. By combining the predictions of multiple models, Stacking mitigates the biases and variances inherent in individual models, resulting in more accurate and reliable predictions. However, it is important to carefully consider certain aspects, including appropriate model selection, thoughtful feature engineering, and the complexity associated with model fusion.

## III. EXPERIMENTAL STUDY

*A. Data Preprocessing and Augmentation*

The dataset used in this study comprises images sourced from the DDSM and CBIS-DDSM datasets. It consists of a total of 55,890 data points, each with two associated labels: a binary label (0 for negative and 1 for positive) and a multi-class label (0 for negative, 1 for benign calcification, 2 for benign mass, 3 for malignant calcification, and 4 for malignant mass). Due to the common occurrence of class imbalance in medical datasets, appropriate data preprocessing and augmentation techniques were employed to enhance the performance of the models. These approaches aimed to address the class imbalance issue, improve the distribution of the data, and increase the robustness of the models for accurate classification results.

In this experiment, negative (DDSM) images were segmented into 598x598 pixel tiles and resized to 224x224 pixels. Positive (CBIS-DDSM) images had ROIs extracted using masks, with added padding for context. Each ROI underwent random cropping three times to create 598x598 pixel crops with random flips and rotations. The images were then resized to 224x224 pixels. This resizing facilitates comparison with models like VGG16 and ResNet-50, which have an input size of 224x224 pixels. Below are sample images from the dataset:

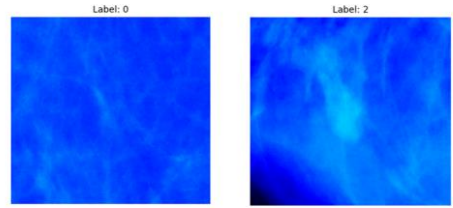

Fig. 2. The images in this experimental dataset have labels of 0 and 2.

*B. Network Achitechure and Implementation*

BSECNN is an ensemble model consisting of multiple CNN networks that utilizes 10 independent sub-models. Each sub-model is individually trained on autonomously generated sub-datasets using bagging with bootstrap sampling. The results of these sub-models are then stacked together to merge their predictions, thereby enhancing the predictive capability of the ensemble model. The stacked meta-model used is a random forest.

In each CNN sub-model, the input layer consists of a 32 3x3 convolutional layer followed by a max-pooling layer. This is followed by 64 and 128 3x3 convolutional layers with max-pooling. The flattened output is then passed through a fully connected layer with 512 neurons, followed by 5 additional fully connected layers.

The ReLU activation function is employed in all layers of this neural network, resulting in a total of 9,683,658 parameters. The architecture of BSECNN is depicted in the diagram below:

```
Layer (type)                    Output Shape              Param #
=================================================================
conv2d (Conv2D)                 (None, 222, 222, 32)      896

max_pooling2d (MaxPooling2D     (None, 111, 111, 32)      0
)

conv2d_1 (Conv2D)               (None, 109, 109, 64)      18496

max_pooling2d_1 (MaxPooling     (None, 54, 54, 64)        0
2D)

conv2d_2 (Conv2D)               (None, 52, 52, 128)       73856

max_pooling2d_2 (MaxPooling     (None, 26, 26, 128)       0
2D)

conv2d_3 (Conv2D)               (None, 24, 24, 128)       147584

max_pooling2d_3 (MaxPooling     (None, 12, 12, 128)       0
2D)

flatten (Flatten)               (None, 18432)             0

dense (Dense)                   (None, 512)               9437696

dense_1 (Dense)                 (None, 10)                5130

=================================================================
Total params: 9,683,658
Trainable params: 9,683,658
Non-trainable params: 0
_________________________________________________________________
```

Fig. 3. Structure of CNN sub-models in the bagging ensemble.

BSECNN integrates the aforementioned 10 sub-models and employs the stacking method to handle the ensemble output of the final model. The stacking meta-model adopts random forest, learning from the outputs of the 10 sub-models to provide the ultimate result of the entire model.

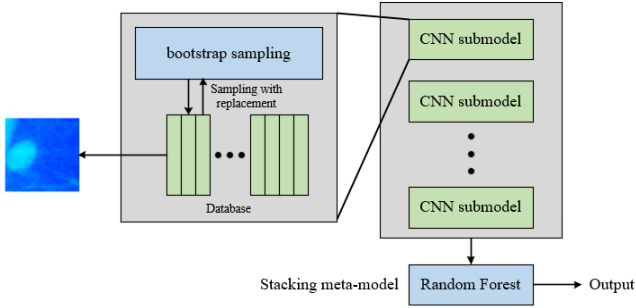

Fig. 4. The structure of the Stacking-Enhanced Bagging Ensemble.

## C. Model Compilation

In this section, we discuss the initialization of hyperparameters, the selection of the loss function, the choice of bagging_ratio and n_models during training of the BSECNN model, as well as the selection of the optimizer.

### 1) Loss Function

In this model, we employ Sparse Categorical Cross-Entropy as the loss function, which is widely used for multi-class classification tasks when the target labels are represented as integers. The Sparse Categorical Cross-Entropy loss function compares the predicted probability distribution with the true class labels. It computes the cross-entropy loss for each sample and computes the average across the entire batch. Unlike Categorical Cross-Entropy, which requires one-hot encoded target labels, Sparse Categorical Cross-Entropy accepts integer labels directly.

For the $i-th$ sample, assuming its true class label is $y_i$ and the predicted class probability distribution is pi, with a total of C classes, the formula for the Sparse Categorical Cross-Entropy loss function is:

$$L = -\sum \log(p_i[y_i]) \quad (5)$$

Here, log represents the natural logarithm, and pi[yi] represents the probability value of the corresponding true class in the predicted probability distribution. The loss function sums the losses for all samples and takes the negative value to obtain the final loss value.

### 2) bagging_ratio and n_models

In this model, there are two important parameters: bagging_ratio and n_models. For the model, a smaller bagging_ratio makes each sub-model more unique, which is advantageous for the stacking ensemble method. However, it also increases the risk of overfitting for the sub-models. On the other hand, a larger bagging_ratio implies a longer training time, but the performance of the sub-models will be better. Additionally, a smaller n_models leads to faster training speed, but the model performance may decrease. Therefore, in this experiment, we tested the accuracy of the model with bagging_ratio values of 0.6, 0.7, and 0.8, and n_models values of 20, 15, and 10. the result is shown in Table 1:

TABLE I. DIFFERENT PARAMETERS FOR ACCURACY

| bagging_ratio | n_models | accuracy |
|---|---|---|
| 0.6 | 20 | 0.9787 |
| **0.7** | **15** | **0.9821** |
| 0.8 | 10 | 0.9815 |

### 3) Optimizer:

In 2014, Kingma and Lei BA [13] proposed the Adam Optimizer, which combines the advantages of the Adagrad and RMSPROP optimization algorithms. Today, Adam is often used as the default optimizer due to its excellent performance.

$$m = \beta_1 \cdot m + (1 - \beta_1) \cdot g \quad (6)$$

$$v = \beta_2 \cdot v + (1 - \beta_2) \cdot g^2 \quad (7)$$

$$\hat{m} = \frac{m}{1 - \beta_1^t} \quad (8)$$

$$\hat{v} = \frac{v}{1 - \beta_2^t} \quad (9)$$

$$\theta = \theta - \frac{\eta}{\sqrt{\hat{v}} + \epsilon} \cdot \hat{m} \quad (10)$$

In this context, we have several parameters involved in the Adam optimizer. θ represents the parameters being optimized, g denotes the gradients, m and v are the exponentially weighted averages of past gradients and squared gradients respectively. The parameters ^m and ^v are the bias-corrected versions of m and v. η is the learning rate, β1 and β2 are the decay rates, and t represents the current iteration or time step. ε is a small constant for numerical stability. These parameters collectively contribute to the efficient and adaptive parameter updates during optimization.

## IV. RESULTS AND DISCUSSIONS

### A. Evaluation metrics

For binary classification problems, we still use accuracy, F1 score, and other metrics. However, for multiclass classification problems, we employ macro-average and micro-average to evaluate the performance of the model. Considering the imbalanced nature of the DDSM dataset, micro-average is more suitable for evaluation due to its better ability to handle imbalanced samples. The formulas for micro-average are as follows:

$$\text{Precision}_{\text{micro}} = \frac{TP_{\text{total}}}{TP_{\text{total}} + FP_{\text{total}}} \quad (11)$$

$$Recall_{micro} = \frac{TP_{total}}{TP_{total} + FN_{total}} \quad (12)$$

$$F1_{micro} = \frac{2 \times Precision_{micro} \times Recall_{micro}}{Precision_{micro} + Recall_{micro}} \quad (13)$$

In these formulas, $TP_{total}$ represents the **total number** of true positives across all classes, $FP_{total}$ represents the total number of false positives across all classes, and $FN_{total}$ represents the total number of false negatives **across all classes**. These formulas are used to calculate the micro-average precision, recall, and $F1$ score, which can be utilized to evaluate the overall performance of a model in the presence of **imbalanced samples**.

*B. Test Result*

*1) Training of the sub-models*

In Figure 5, we observe that the iteration process is mutually stable and converges around the 14th epoch. Some overfitting occurs after 25 epochs, which can be attributed to the lower sampling ratio of the dataset (to expedite model training). The accuracy on the test set is 95.47%, with a test loss of 0.2552. This is just one of the models, and by stacking models, the accuracy can be further improved to 98.34%. These results highlight the excellent performance of BSBCNN in breast cancer classification tasks.

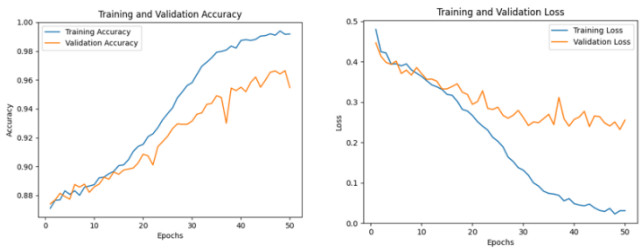

Fig. 5. line graph of the training accuracy and loss for a single bagging submodel, with a ratio of 0.7, n_models of 15, and 50 epochs.

*2) Method of sub-model ensemble*

We have investigated three methods for ensemble of sub-models, namely averaging, voting, and stacking. The stacking method uses random forest as the meta-model. We evaluated the performance of these three methods in terms of micro-precision, micro-recall, and micro-F1 score. Ultimately, we selected the stacking method with random forest as the meta-model. This method not only improves the model's accuracy but also enhances its robustness. The test results are shown in Table 2:

TABLE II. COMPARISON OF MODEL PERFORMANCE FOR DIFFERENT METHODS

| Methods | $Precision_{micro}$ | $Recall_{micro}$ | $F1_{micro}$ |
|---|---|---|---|
| Average | 0.9018 | 0.9308 | 0.9153 |
| Vote | 0.9195 | 0.9414 | 0.9301 |
| **stacking** | **0.9358** | **0.9480** | **0.9419** |

The table shows that there is not much difference in performance between the average and voting methods, but the stacking method performs better. Moreover, the metrics of micro-precision, micro-recall, and micro-F1 score provide better evaluation for imbalanced datasets.

*3) Overall Model Performance*

Analysis of the overall model performance is mainly done using the model's confusion matrix, which provides a visual representation of the model's classification results for the dataset. The confusion matrix for the 5-class model is as follows:

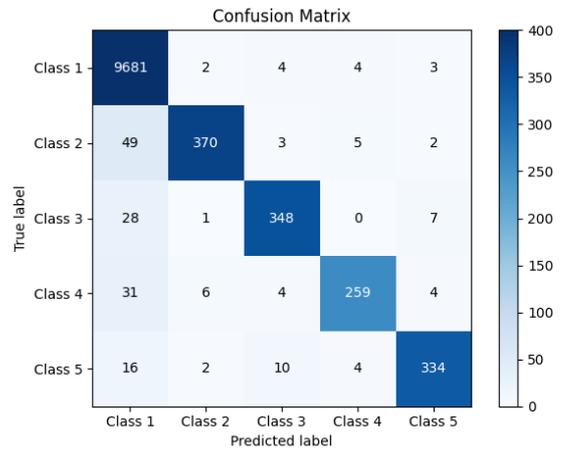

Fig. 6. Confusion matrix for BSECNN on the DDSM public dataset.

From Figure 6, it can be seen that the model's performance is very impressive, with an accuracy of 98.34% at this point.

*C. Comparison experiments*

In the comparative experiments, we evaluated the performance of three different models, namely BSECNN, VGG16, and ResNet-50, on the same dataset.

VGG16 is a convolutional neural network architecture developed by the Visual Geometry Group at the University of Oxford. It was introduced in 2014 by Karen Simonyan and Andrew Zisserman in their paper [14]. One notable characteristic of VGG16 is its deep stacking of convolutional layers, allowing the network to learn features at various abstract levels. This deep architecture enables the model to capture intricate patterns and details in images, making it particularly suitable for tasks like object recognition and localization.

On the other hand, ResNet-50 is a deep residual network architecture proposed by Microsoft Research and introduced in a paper in 2015 [15]. It belongs to the ResNet series of models and represents a significant breakthrough in addressing the challenges of gradient vanishing and exploding in deep networks. The residual connections in ResNet-50 enable the network to learn residual mappings, which alleviate the degradation problem and facilitate the training of very deep models.

In the comparative experiment on the DDSM dataset, we assessed the performance of BSECNN, VGG16, and ResNet-50. The results of this evaluation are presented in the table below:

TABLE III. COMPARSION OF ACCURACY FOR DIFFERENT MODELS

| Models | Accuracy |
|---|---|
| VGG16 | 0.9012 |
| ResNet-50 | 0.9201 |
| **BSECNN** | **0.9834** |

From Table 3, it is evident that BSECNN outperforms both VGG16 and ResNet-50 in the task of breast cancer classification.

## V. CONCLUSION

In this study, we propose a convolutional neural network (CNN) classification network for breast cancer classification, based on the Bagging and Stacked Ensemble Learning methods. The model is trained and tested on the DDSNN public dataset, and it can classify input images quickly and accurately. According to our research results, for binary classification (presence or absence of breast cancer), the accuracy reaches 98.84%, while for five-class classification, the accuracy reaches 98.34%. Furthermore, the model achieves a micro-average recall of 94.80% and an F1 score of 94.19%. We conducted comparative experiments to examine the influence of different bagging_ratio and n_models values on the model, as well as several ensemble bagging methods. It is worth noting that under the same parameter settings, our BSECNN surpasses VGG16 and ResNet-50 in terms of accuracy, with an improvement of 8.22% and 6.33% respectively.

## ACKNOWLEDGMENT

Peicheng Wu and Runze Ma extend their heartfelt gratitude to Prof. Teoh Teik Toe, Director of the AI Lab at Nanyang Technological University, for his invaluable mentorship and significant contributions to their research in image recognition. Prof. Teoh Teik Toe's profound expertise and insightful guidance have been instrumental in honing their work on image classification and detection. His assistance throughout the preparation of this thesis has not only enhanced the effectiveness of their research but also fostered their academic growth. Without his steadfast support and dedication, the advancement of their work would not have been conceivable. They express their deepest appreciation for his unwavering commitment and exemplary mentorship.

## REFERENCES

[1] " Subtypes of Breast Cancer: A Review for Breast Radiologists." Journal of Breast Imaging (2020): n. pag.

[2] Chaturvedi, Aditi and Bhawna Sirohi. "Classification of Breast Cancer." Breast Cancer (2022): n. pag.

[3] " Orrantia-Borunda, Erasmo, et al. "Subtypes of Breast Cancer." Breast Cancer, edited by Harvey N. Mayrovitz, Exon Publications, 6 August 2022. doi:10.36255/exon-publications-breast-cancer-subtypes.

[4] Łukasiewicz, Sergiusz, et al. "Breast Cancer—Epidemiology, Risk Factors, Classification, Prognostic Markers, and Current Treatment Strategies—An Updated Review." Cancers, vol. 13, no. 17, Aug. 2021, p.4287. Crossref.

[5] Harbeck, N., Penault-Llorca, F., Cortes, J. et al. Breast cancer. Nat Rev Dis Primers 5, 66 (2019).

[6] Breiman, L.. "Bagging predictors." Machine Learning 24 (2004): 123-140.

[7] Heath, Michael D. et al. "THE DIGITAL DATABASE FOR SCREENING MAMMOGRAPHY." (2007).

[8] Heath, Michael D. et al. "Current Status of the Digital Database for Screening Mammography." Digital Mammography / IWDM (1998).

[9] Lévy, Daniel and Arzav Jain. "Breast Mass Classification from Mammograms using Deep Convolutional Neural Networks." ArXiv abs/1612.00542 (2016): n. pag.

[10] L. Breiman, "Bagging Predictors," in Machine Learning, vol. 24, no. 2, pp. 123-140, 1996.

[11] B. Efron, "Bootstrap methods: Another look at the jackknife," The Annals of Statistics, vol. 7, no. 1, pp. 1-26, 1979

[12] D. H. Wolpert, "Stacked generalization," Neural Networks, vol. 5, no. 2, pp. 241-259, 1992

[13] D. Kingma and J. Ba, "Adam: A Method for Stochastic Optimization," Computer Science, 2014.

[14] K. Simonyan and A. Zisserman, "Very Deep Convolutional Networks for Large-Scale Image Recognition," in 3rd International Conference on Learning Representations (ICLR), San Diego, CA, USA, May 2015.

[15] K. He, X. Zhang, S. Ren, and J. Sun, "Deep Residual Learning for Image Recognition," in IEEE Conference on Computer Vision and Pattern Recognition (CVPR), Las Vegas, NV, USA, Jun. 2016.